\documentclass[a4wide]{article}

\usepackage{textcomp}
\usepackage[utf8]{inputenc}
\usepackage[T1]{fontenc}

\usepackage[round]{natbib}

\usepackage{hyperref}
\usepackage[dvipsnames]{xcolor}

\usepackage{amssymb}
\usepackage{amsmath}
\usepackage{amsthm}
\usepackage{mathtools}
\usepackage{algorithm,algorithmic}
\usepackage{stmaryrd}
\usepackage{booktabs}
\usepackage{color}
\usepackage{nicefrac}
\usepackage{multirow}
\usepackage{lipsum}
\usepackage{graphicx}
\usepackage{subcaption}
\usepackage{microtype}

\begin{document}

\begin{center}
 {\Large Beyond Static Instruction: A Multi-agent AI Framework for Adaptive Augmented Reality Robot Training}
\end{center}

\vspace{7mm}

\noindent\textbf{Nicolas Leins}\hfill\href{mailto:leins@zib.de}{\ttfamily leins@zib.de}\\
\emph{\small Zuse Institute Berlin \\
Berlin, Germany}\\
\\
\textbf{Jana Gonnermann-Müller}\hfill\href{mailto:gonnermann-mueller@zib.de}{\ttfamily gonnermann-mueller@zib.de}\\
\emph{\small Zuse Institute Berlin \& Weizenbaum Institute Berlin\\
Berlin, Germany}\\
\\
\textbf{Malte Teichmann}\hfill\href{mailto:malte.teichmann@wi.uni-potsdam.de}{\ttfamily malte.teichmann@wi.uni-potsdam.de}\\
\emph{\small University of Potsdam\\
Potsdam, Brandenburg, Germany\\
Weizenbaum Institute Berlin\\ 
Berlin, Germany}\\
\\
\textbf{Sebastian Pokutta}\hfill\href{mailto:pokutta@zib.de}{\ttfamily pokutta@zib.de}\\
\emph{\small Zuse Institute Berlin \& TU Berlin\\
Berlin, Germany}\\

\vspace{5mm}

\begin{center}
\begin{minipage}{0.85\textwidth}
\begin{center}
 \textbf{Abstract}
\end{center}
 {\small Augmented Reality (AR) offers powerful visualization capabilities for industrial robot training, yet current interfaces remain predominantly static, failing to account for learners' diverse cognitive profiles. In this paper, we present an AR application for robot training and propose a multi-agent AI framework for future integration that bridges the gap between static visualization and pedagogical intelligence. We report on the evaluation of the baseline AR interface with 36 participants performing a robotic pick-and-place task. While overall usability was high, notable disparities in task duration and learner characteristics highlighted the necessity for dynamic adaptation. To address this, we propose a multi-agent framework that orchestrates multiple components to perform complex preprocessing of multimodal inputs (e.g., voice, physiology, robot data) and adapt the AR application to the learner's needs. By utilizing autonomous Large Language Model (LLM) agents, the proposed system would dynamically adapt the learning environment based on advanced LLM reasoning in real-time.}
\end{minipage}
\end{center}

\vspace{0mm}

\section{Introduction} \label{sec:introduction}

As robotic systems become increasingly complex and ubiquitous in industrial settings, the operational bottleneck is no longer the hardware capabilities but the human operator's ability to interact with them efficiently. Controlling and programming a robotic arm remains a cognitively demanding task, particularly for novices. Consequently, the effective training and onboarding of new workers is crucial for operational success. However, for years, vocational training in this domain has relied on static resources like manuals, video tutorials, or rigid step-by-step instructions. Beyond the inherent complexity of the task itself, an additional barrier stems from the industry's continued dependence on 2D interaction interfaces. These interfaces introduce extra cognitive load, requiring users to mentally project complex, spatial robot motions onto 2D displays.

To better convey the 3D and visually complex nature of robot motion, Augmented Reality (AR) has emerged as an effective solution \cite{chang_survey_2024}. Because robotic manipulation inherently involves 3D spatial knowledge and relations, the nature of the task implies the use of a spatial interface. AR addresses this by facilitating interaction with 3D content contextually embedded in the real surroundings, removing the need for 2D abstraction \cite{azuma_survey_1997}. Research has shown that this spatial alignment is helpful in significantly reducing cognitive load compared to traditional interfaces \cite{howard_meta-analysis_2023, yang_har2bot_2024}. AR interfaces for HRI have also demonstrated increased usability \cite{tsamis_intuitive_2021}, superior spatial understanding \cite{kumar_mixed_2025}, and reduced complexity \cite{ong_augmented_2020, neves_application_2020, ikeda_programar_2024, lotsaris_ar_2021} compared to conventional interfaces and interaction techniques.

Yet, despite these visual and spatial advantages, most current AR interfaces remain 'static' in terms of the information and instruction they provide. These 'one-size-fits-all' approaches fail to account for the heterogeneity of learners. For example, they display the same visualizations and instruction texts regardless of the user's proficiency or stress level. Ideally, an AR learning application to teach robot interaction has a detailed understanding of the learner and can adapt to meet the learner's needs \cite{gonnermann-muller_unlocking_2025}.

\begin{figure}[ht]
  \centering
  \includegraphics[alt={Multi-agent framework diagram and AR application interface showing input layer with speech, task, robot and physiological analysis, reasoning layer with assessment and teacher agents, and output layer with tutor, visualization and instruction agents},width=\textwidth]{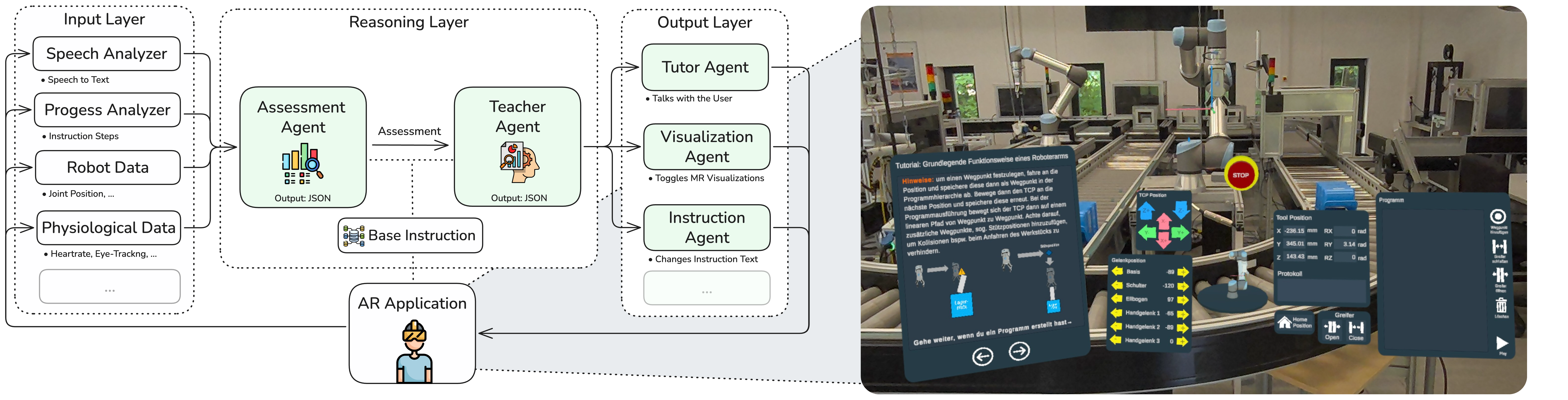}
  \caption{Multi-agent framework visualization and screenshot of AR application user interface}
  \label{fig:multi-agent-framework}
\end{figure}

To address this, we propose an AI-powered framework (\autoref{fig:multi-agent-framework}) that combines immersive AR with the adaptive capabilities of Large Language Models (LLMs). While LLMs mimic human-like reasoning \cite{yao_tree_2023}, they are prone to hallucination and context overflow \cite{guo_large_2024}. To mitigate these risks, we propose a multi-agent framework where specialized agents cooperatively manage sensory processing, pedagogical reasoning, and instruction. Section \ref{sec:multi-agent-framework} details this architecture.
Our objective is to create an AR system that dynamically adapts to individual needs. In this paper, we present a fully implemented AR application for robotic training and a conceptual multi-agent architecture for adaptive support. We describe the system development, preliminary evaluation, and the roadmap for integrating the proposed adaptive framework.

\section{System Architecture}

To realize a fully adaptive learning environment, we have developed a modular architecture comprising a foundational AR interface and propose a multi-agent backend for future integration. In this section, we describe our AR application, and present a preliminary user evaluation to ensure basic usability. Subsection \ref{sec:multi-agent-framework} covers the proposed architecture of the multi-agent framework, which represents a conceptual design for future implementation.

\subsection{Augmented Reality Application}

To create the immersive AR learning environment, we developed an application using Unity (2022.3) and deployed it on the Meta Quest 3. This video see-through (VST) head-mounted display (HMD) was selected for its high-resolution passthrough (18 PPD) and wide field of view, ensuring clear visibility of the physical robotic arm while overlaying virtual content.
The complete application is available in a public repository \url{https://github.com/NLeins/UR5e-Augmented-Reality-Quest3-Interface}.

\paragraph{Interaction Design} The AR application's UI replicates the fundamental functionality of a standard industry teach pendant. Interaction is achieved via bare-hand tracking. Users press virtual buttons with their index fingers to control the robot, eliminating the need for handheld controllers. The application supports both joint-based and translational Tool Center Point (TCP) movement, as well as standard waypoint programming logic.

\paragraph{Data Synchronization} We utilized the Robot Operating System 2 (ROS2) as the middleware to establish bidirectional communication between the Unity application and a Universal Robots UR5e robotic arm \cite{noauthor_ur5e_2024}. Using the Unity-ROS-TCP connector \cite{unity_technologies_ros_2025}, the application subscribes to the robot's joint states to mirror its pose in real-time and publishes URScript commands to control the physical arm in external control mode. Spatial alignment between virtual objects and the robot is maintained via a persistent spatial anchor at the robot's base.

\begin{figure}[ht]
    \centering
    \begin{subfigure}{0.4\columnwidth}
        \centering
        \includegraphics[alt={AR application screenshot showing a robotic arm with colored text labels identifying joint positions},height=2.2in]{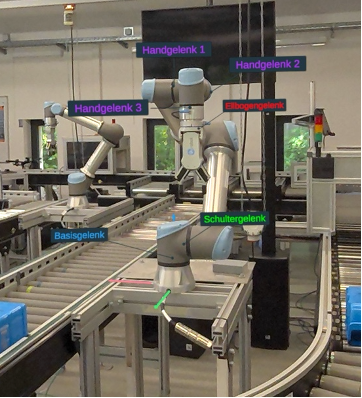}
        \caption{Joint labels}
        \label{fig:joint_labels}
    \end{subfigure}%
    \begin{subfigure}{0.4\columnwidth}
        \centering
        \includegraphics[alt={AR coordinate systems superimposed on industrial robotic arm showing red x-axis, green y-axis, and blue z-axis},height=2.2in]{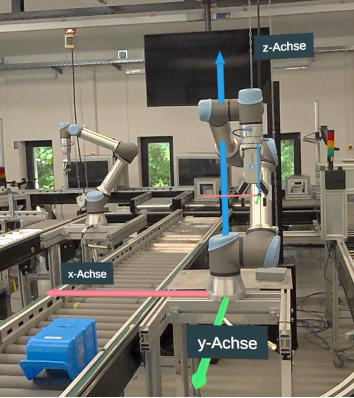}
        \caption{Base coordinate system}
        \label{fig:coordinates}
    \end{subfigure}\\[0.5cm]
    \begin{subfigure}{\columnwidth}
        \centering
        \includegraphics[alt={AR trajectory visualization showing five blue diamond waypoint markers connected by lines representing the robot's programmed path},height=2.5in]{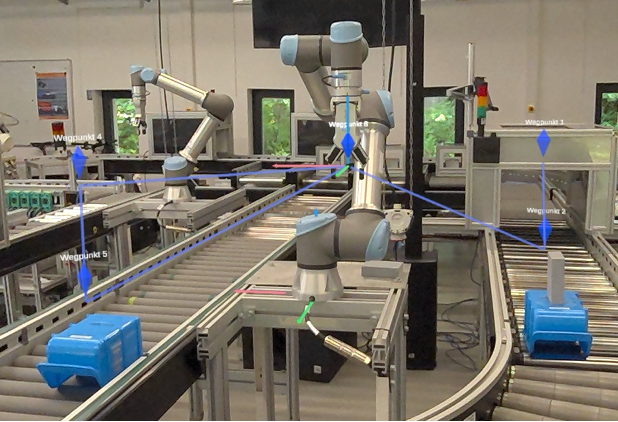}
        \caption{Waypoints and execution path}
        \label{fig:waypoints}
    \end{subfigure}
    \caption{Visualizations in the AR Application}
    \label{fig:AR-visualizations}
\end{figure}

\paragraph{Spatial Visualizations} Key features include: (1) the robot's coordinate system visualized at the TCP, allowing users to map intended directions without complex mental rotations \autoref{fig:coordinates}, (2) spatial waypoint markers connected by path lines for trajectory visualization and debugging \autoref{fig:waypoints}, and (3) dynamic tooltips on the physical robot corresponding to the current learning step \autoref{fig:joint_labels}.

\subsection{Preliminary Evaluation}

Before implementing the multi-agent framework into the AR application to enable adaptation, we first evaluated the base version to ensure its suitability. Therefore, we designed a standardized learning task teaching the fundamentals of robot control to novices. The task consisted of three topics. It started with basic joint control. Users learned how to maneuver the robot to a target pose using individual joint rotations. After that, users learned to use the linear displacement of the TCP in Cartesian space. To reinforce understanding, they were tasked with reaching the same target pose as in Step 1 but using translational inputs. The final task required users to program a whole sequence: moving the robot to a starting position, picking up a workpiece using the mounted OnRobot RG2 \cite{noauthor_rg2_2025} gripper, and placing it on a target carrier. This required defining waypoints, utilizing intermediate support positions to avoid collisions, and executing the program logic.

To validate the baseline AR interaction and identify requirements for the adaptive framework, we conducted a user study with $N=36$ participants. The sample consisted of 27 men and nine women with an average age of 27 $(SD=6.7)$. Following the learning task, participants completed a questionnaire assessing system usability using the \textit{System Usability Scale} (SUS) \cite{brooke_sus_1996}, extraneous cognitive load (ECL) using the naive rating scale from Klepsch et al. \cite{klepsch_development_2017} and learner characteristics including the \textit{Mental Rotation Test} (MRT) \cite{peters_redrawn_1995}, Affinity for Technology Interaction (ATI) \cite{franke_personal_2018} and prior experience in robotics (ER). Participants were post-hoc categorized into 'High' and 'Low' groups for MRT, ATI, and ER based on the sample mean.

The system achieved an overall mean SUS score of 82.6 $(SD=14.1)$, corresponding to a very high rating, with a low overall ECL $(M=1.70, SD=0.7)$, indicating overall suitability for supporting the task and the appropriate design of the learning material.

Nevertheless, the user study revealed differences among participants regarding the required time to complete the task. The task duration ranged widely from 14 to 33 minutes $(M=23.1, SD=4.8)$. In terms of prior knowledge about robotics, High-Experience users reported lower ECL ($M=1.56$) and higher SUS ($M=89.2$) compared to Low-Experience users $(ECL: M=1.78, SUS: M=78.8)$. A similar pattern emerged for spatial ability. High-MRT users reported lower ECL ($M=1.55$) and higher SUS ($M=85.1$) compared to Low-MRT users, who showed higher ECL $(M=1.84)$ and lower SUS ($M=80.3$). Users with low ATI rated SUS nearly 13 points lower $(M=75.8)$ than High-ATI users ($M=88$) and reported higher ECL ($M_{low ATI}=1.85, M_{high ATI}= 1.58$).

These observed patterns suggest that 'one size does not fit all'. Users with lower spatial ability, a lower ATI, or less experience tended to perceive the static interface as more burdensome (higher ECL) and less usable (lower SUS). This provides motivation for implementing our adaptive multi-agent framework. An adaptive system could specifically target these 'Low' groups to reduce cognitive load, while fading these aids for capable users to maintain flow. Building on these findings, we next introduce our conceptual framework designed to operationalize such adaptation.

\subsection{Multi-Agent AI Framework}
\label{sec:multi-agent-framework}

To bridge the gap between the static AR interface and the diverse user needs identified in our evaluation, we propose a conceptual multi-agent framework for future implementation. While the AR application described above is fully implemented and available, the multi-agent AI integration framework presented in this section represents a proposed architecture for future development. This architecture is grounded in the distinction between personalization (tailoring the system to static user traits, e.g., low spatial ability) and adaptation (dynamic runtime adjustments). Our goal is to develop a system that evolves with the learner over time. To do so, we ground the design of the learning environment in established instructional frameworks, including Mayer's Multimedia Principles \cite{mayer_multimedia_2020} and cognitive load theory \cite{sweller_cognitive_2019}. These principles guide how information is presented, structured, and paced to support effective learning. As a result, the proposed system would adapt not only what it shows but also how it shows it. For example, as a novice user gains confidence, the interface could gradually transition from high-scaffolding, explanatory guidance toward more expert-oriented support.

To achieve this, we propose treating the 'Teacher' not as a monolithic algorithm but as a group of agents that would process multimodal data, including text, voice, gaze, and physiological metrics, to modify the AR application in real-time. The proposed architecture is designed with a layered communication structure, where the AR application would serve as both the sensor platform and the actuator interface. The backend logic would be structured into a hierarchical pipeline comprising three distinct layers: a deterministic \textit{Input Layer} for sensory processing, a \textit{Reasoning Layer} driven by LLMs for pedagogical decision-making, and an \textit{Output Layer} for executing specific interventions.

\paragraph{Input Layer}

The proposed \textit{Input Layer} would serve as the direct interface to the AR application and hardware sensors. This layer would primarily consist of deterministic modules. We propose implementing these as deterministic modules to ensure that the data fed to the \textit{Reasoning Layer} is grounded in ground-truth sensor readings, preventing early-stage hallucination. These modules would act as gateways to preprocess the high-frequency, unstructured data generated by the AR hardware and the robot. Specific modules would include a \textit{Voice Analyzer} for speech-to-text conversion, a \textit{Progress Analyzer} to track the current instructional step by monitoring state transitions in the AR application, a \textit{Robot Data Analyzer} to monitor joint states and derive kinematic metrics (e.g., velocity, acceleration), and a \textit{Physiological Analyzer} that would combine measures like heart rate variability (HRV) from wearable sensors to derive cognitive factors such as stress or attention. This layer would be designed to be easily extended with additional modules for other data sources if the use case requires it. The primary function of this layer would be to prepare semantic data. For instance, converting raw eye-tracking coordinates into meaningful semantic events, such as 'user is fixating on the gripper', using gaze fixation algorithms. These modules would parse the vast stream of raw inputs into a predefined, structured output format (e.g., JSON schema) that serves as the ground truth for the \textit{Reasoning Layer}.

\paragraph{Reasoning Layer}

The core intelligence would reside in the proposed \textit{Reasoning Layer}. To prevent context overflow and maintain the separation of subtasks, we propose decoupling the task of 'Understanding' from 'Deciding' by employing two specialized LLM agents. First, the \textit{Assessment Agent} would receive the formatted data from the \textit{Input Layer}. Its system prompt would restrict it to synthesizing an overall assessment of the user's current situation. E.g., if the \textit{Input Layer} reports that the user asked 'I don't get this', their heart rate is elevated, and the instruction is at step four, the \textit{Assessment Agent} would synthesize this into a condensed, semantically enriched text describing the user as frustrated with the content or task in step 4. The agent would maintain a temporal context window (e.g., last 30 seconds of interaction) to track state transitions and avoid oscillating assessments. This assessment would be passed to the \textit{Teacher Agent}. Without the need to analyze raw data, the \textit{Teacher Agent} would focus entirely on pedagogical strategy. It would reason based on the \textit{Assessment Agent's} summary and a shared knowledge base containing instructional content, task information, and pedagogical rules. The agent would determine the optimal intervention, for instance, determining that the user requires emotional encouragement rather than technical correction. After this decision, it would delegate tasks to the appropriate agents in the \textit{Output Layer}. Both agents in the \textit{Reasoning Layer} would be configured with high temperature settings to enable creative reasoning about application adaptation strategies.

\paragraph{Output Layer}

The proposed \textit{Output Layer} would consist of specialized agents that could execute specific changes within the AR application. These agents, also powered by LLMs, would receive high-level calls to action from the \textit{Reasoning Layer} and translate them into machine-readable formats executable by the program logic. In contrast to the \textit{Reasoning Layer}, agents in the \textit{Output Layer} would use lower temperature settings and be explicitly instructed to adhere strictly to predefined JSON schemas, generating structured data for machine interpretability. For example, the \textit{Tutor Agent} could be instructed to mimic a human tutor, generating empathetic, human-like text to be spoken by a virtual avatar. The \textit{Visualization Agent}, having an overview of available graphical assets, could reason about spatial interventions, such as generating an additional arrow to guide a user struggling with mental rotation, with parameters for position, scale, and color. Finally, the \textit{Instruction Agent} would generate adapted textual content, e.g., rewriting complex explanations into simpler language. Communication between the agents and the AR application would be implemented by outputting predefined JSON files that are sent to the AR application. The AR application would parse these JSON files to call local functions, which would then execute the corresponding adaptations within the application. By keeping these agents specialized and constraining their output to predefined JSON schemas, the proposed framework would maintain a controlled, well-defined behavior structure while preserving the advanced, human-like ability to reason creatively about how to best support the learner.

This proposed design would enforce a rigid architectural structure with strict semantic separation of tasks, while allowing the individual agents within these layers to operate with significant autonomy. Notably, the LLM-based agents would not be designed as simple pass-through functions. Instead, they would independently evaluate the necessity and nature of their responses. For instance, the \textit{Assessment Agent} would evaluate incoming data against context history and may decide that minor data fluctuations do not warrant a revision of the user's state, effectively filtering out noise. Similarly, the \textit{Teacher Agent} would autonomously determine \textit{if} an intervention is necessary and \textit{which} output agent is best suited to execute it, preventing unnecessary disruptions. This combination would ensure the system maintains structural stability while leveraging the nuanced, human-like reasoning required for effective teaching. Collectively, these layers would establish a closed-loop adaptive system that continuously cycles through sensing (Input), understanding (Reasoning), and acting (Output). This architecture aims to transform the AR interface from a passive visualization tool into an active pedagogical partner, capable of delivering personalized support strategies necessitated by the variance observed in our user study.

\subsection{Ethical Considerations and Limitations}

We address biometric data privacy through three strategies: First, deterministic modules in the Input Layer act as gatekeepers, converting raw sensors (including gaze) into abstract semantic events before they reach any LLM, ensuring transparency. Second, the modular design allows users to disable specific sensors. Third, by decomposing reasoning into specialized sub-agents, the architecture is optimized for small, locally hosted models, preventing sensitive data from leaving the user's environment.
A remaining challenge is the non-deterministic nature of LLMs, which may lead to inconsistent pedagogical decisions. While our multi-agent architecture introduces checkpoints (e.g., the Teacher Agent validating the Assessment Agent's output), these safeguards cannot fully eliminate unpredictability. Future work must rigorously evaluate the system's reliability and its impact on learning outcomes.

\section{Conclusion}

We presented a fully implemented, open-source AR robot training application and a conceptual multi-agent framework for dynamic adaptation. While our baseline AR system demonstrated high usability, evaluation revealed that diverse learner needs require adaptive support. Our proposed framework addresses this by decoupling sensory processing from pedagogical reasoning using specialized LLM agents. Future work will focus on integrating this framework and evaluating its impact on learning outcomes in a comparative study.

\section{Acknowledgments}

This research was funded by the German Federal Ministry of Research, Technology and Space (BMFTR) under the project ‘German Internet Institute’ (16DII147) and fund number 16IS23025B; by the DFG Cluster of Excellence MATH+ (EXC-2046/1, project id 390685689) funded by the German Research Foundation (DFG); and through the Research Campus Modal funded by the German Federal Ministry of Education and Research (fund numbers 05M14ZAM, 05M20ZBM). The authors thank the Chair of Business Informatics, Processes and Systems at the University of Potsdam for providing the technical hardware used in this experiment.

\clearpage
\appendix
\onecolumn

\bibliographystyle{ACM-Reference-Format}
\bibliography{references_zotero}

@article{howard_meta-analysis_2023,
	title = {A {Meta}-analysis of augmented reality programs for education and training},
	volume = {27},
	issn = {1434-9957},
	url = {https://doi.org/10.1007/s10055-023-00844-6},
	doi = {10.1007/s10055-023-00844-6},
	journal = {Virtual Reality},
	author = {Howard, Matt C. and Davis, Maggie M.},
	month = aug,
	year = {2023},
	pages = {2871--2894},
}

@book{mayer_multimedia_2020,
	address = {Cambridge},
	edition = {3},
	title = {Multimedia {Learning}},
	doi = {10.1017/9781316941355},
	publisher = {Cambridge University Press},
	author = {Mayer, Richard E.},
	year = {2020},
}

@misc{noauthor_rg2_2025,
	title = {{RG2} {Gripper} - {Flexible} 2 {Finger} {Robot} {Gripper}},
	url = {https://onrobot.com/us/products/rg2-gripper},
	language = {en-US},
	urldate = {2025-07-11},
	journal = {OnRobot},
	year = {2025},
}

@misc{noauthor_ur5e_2024,
	title = {{UR5e} {Lightweight}, versatile cobot},
	url = {https://www.universal-robots.com/products/ur5e/},
	urldate = {2024-12-12},
	year = {2024},
}

@misc{unity_technologies_ros_2025,
	title = {{ROS} {TCP} {Endpoint}},
	copyright = {Apache-2.0},
	url = {https://github.com/Unity-Technologies/ROS-TCP-Endpoint},
	urldate = {2025-07-03},
	publisher = {GitHub},
	author = {{Unity Technologies}},
	year = {2025},
}

@inproceedings{yao_tree_2023,
	address = {Red Hook, NY, USA},
	series = {{NIPS} '23},
	title = {Tree of thoughts: deliberate problem solving with large language models},
	shorttitle = {Tree of thoughts},
	urldate = {2025-12-03},
	booktitle = {Proceedings of the 37th {International} {Conference} on {Neural} {Information} {Processing} {Systems}},
	publisher = {Curran Associates Inc.},
	author = {Yao, Shunyu and Yu, Dian and Zhao, Jeffrey and Shafran, Izhak and Griffiths, Thomas L. and Cao, Yuan and Narasimhan, Karthik},
	year = {2023},
	pages = {11809--11822},
}

@article{chang_survey_2024,
	title = {A {Survey} of {Augmented} {Reality} for {Human}–{Robot} {Collaboration}},
	volume = {12},
	copyright = {http://creativecommons.org/licenses/by/3.0/},
	issn = {2075-1702},
	url = {https://www.mdpi.com/2075-1702/12/8/540},
	doi = {10.3390/machines12080540},
	language = {en},
	number = {8},
	urldate = {2024-09-26},
	journal = {Machines},
	author = {Chang, Christine T. and Hayes, Bradley},
	month = aug,
	year = {2024},
	pages = {540},
}

@article{franke_personal_2018,
	title = {A {Personal} {Resource} for {Technology} {Interaction}: {Development} and {Validation} of the {Affinity} for {Technology} {Interaction} ({ATI}) {Scale}},
	volume = {35(6)},
	doi = {10.1080/10447318.2018.1456150},
	journal = {International Journal of Human–Computer Interaction},
	author = {Franke, Thomas and Attig, Christiane and Wessel, Daniel},
	year = {2018},
	pages = {456--467},
}

@article{peters_redrawn_1995,
	title = {A {Redrawn} {Vandenberg} and {Kuse} {Mental} {Rotations} {Test} - {Different} {Versions} and {Factors} {That} {Affect} {Performance}},
	volume = {28},
	issn = {0278-2626},
	url = {https://www.sciencedirect.com/science/article/pii/S0278262685710329},
	doi = {10.1006/brcg.1995.1032},
	number = {1},
	urldate = {2024-10-08},
	journal = {Brain and Cognition},
	author = {Peters, M. and Laeng, B. and Latham, K. and Jackson, M. and Zaiyouna, R. and Richardson, C.},
	month = jun,
	year = {1995},
	pages = {39--58},
}

@article{azuma_survey_1997,
	title = {A {Survey} of {Augmented} {Reality}},
	volume = {6 (4)},
	doi = {10.1162/pres.1997.6.4.355},
	language = {en},
	journal = {Presence: Teleoperators and Virtual Environments},
	author = {Azuma, Ronald T},
	year = {1997},
	pages = {355--385},
}

@article{gonnermann-muller_unlocking_2025,
	title = {Unlocking {Augmented} {Reality} {Learning} {Design} {Based} on {Evidence} {From} {Empirical} {Cognitive} {Load} {Studies} - {A} {Systematic} {Literature} {Review}},
	volume = {41},
	issn = {1365-2729},
	url = {https://onlinelibrary.wiley.com/doi/abs/10.1111/jcal.13095},
	doi = {10.1111/jcal.13095},
	language = {en},
	number = {1},
	urldate = {2024-12-04},
	journal = {Journal of Computer Assisted Learning},
	author = {Gonnermann-Müller, Jana and Krüger, Jule M.},
	year = {2025},
	note = {\_eprint: https://onlinelibrary.wiley.com/doi/pdf/10.1111/jcal.13095},
	pages = {e13095},
}

@incollection{brooke_sus_1996,
	address = {London},
	title = {{SUS}: {A} '{Quick} and {Dirty}' {Usability} {Scale}},
	isbn = {978-0-429-15701-1},
	url = {https://doi.org/10.1201/9781498710411-35},
	booktitle = {Usability {Evaluation} {In} {Industry}},
	publisher = {CRC Press},
	author = {Brooke, John},
	editor = {Jorden, P. W. and Thomas, B. and Weerdmeester, B. A. and McClelland, I. L.},
	year = {1996},
	pages = {189--194},
}

@article{klepsch_development_2017,
	title = {Development and {Validation} of {Two} {Instruments} {Measuring} {Intrinsic}, {Extraneous}, and {Germane} {Cognitive} {Load}},
	volume = {8},
	issn = {1664-1078},
	url = {https://www.frontiersin.org/journals/psychology/articles/10.3389/fpsyg.2017.01997/full},
	doi = {10.3389/fpsyg.2017.01997},
	language = {English},
	urldate = {2024-11-06},
	journal = {Frontiers in Psychology},
	publisher = {Frontiers},
	author = {Klepsch, Melina and Schmitz, Florian and Seufert, Tina},
	month = nov,
	year = {2017},
}

@article{sweller_cognitive_2019,
	title = {Cognitive {Architecture} and {Instructional} {Design}: 20 {Years} {Later}},
	volume = {31},
	issn = {1040-726X, 1573-336X},
	shorttitle = {Cognitive {Architecture} and {Instructional} {Design}},
	url = {http://link.springer.com/10.1007/s10648-019-09465-5},
	doi = {10.1007/s10648-019-09465-5},
	language = {en},
	number = {2},
	urldate = {2022-01-20},
	journal = {Educational Psychology Review},
	author = {Sweller, John and van Merriënboer, Jeroen J. G. and Paas, Fred},
	month = jun,
	year = {2019},
	pages = {261--292},
}

@article{lotsaris_ar_2021,
	series = {8th {CIRP} {Conference} of {Assembly} {Technology} and {Systems}},
	title = {{AR} based robot programming using teaching by demonstration techniques},
	volume = {97},
	issn = {2212-8271},
	url = {https://www.sciencedirect.com/science/article/pii/S2212827120314906},
	doi = {10.1016/j.procir.2020.09.186},
	urldate = {2024-09-27},
	journal = {Procedia CIRP},
	author = {Lotsaris, Konstantinos and Gkournelos, Christos and Fousekis, Nikos and Kousi, Niki and Makris, Sotiris},
	month = jan,
	year = {2021},
	pages = {459--463},
}

@article{ikeda_programar_2024,
	title = {{PRogramAR}: {Augmented} {Reality} {End}-{User} {Robot} {Programming}},
	volume = {13},
	shorttitle = {{PRogramAR}},
	url = {https://dl.acm.org/doi/10.1145/3640008},
	doi = {10.1145/3640008},
	number = {1},
	urldate = {2024-09-26},
	journal = {J. Hum.-Robot Interact.},
	author = {Ikeda, Bryce and Szafir, Daniel},
	year = {2024},
	pages = {1--20},
}

@inproceedings{tsamis_intuitive_2021,
	title = {Intuitive and {Safe} {Interaction} in {Multi}-{User} {Human} {Robot} {Collaboration} {Environments} through {Augmented} {Reality} {Displays}},
	issn = {1944-9437},
	url = {https://ieeexplore.ieee.org/document/9515474},
	doi = {10.1109/RO-MAN50785.2021.9515474},
	urldate = {2025-06-27},
	booktitle = {2021 30th {IEEE} {International} {Conference} on {Robot} \& {Human} {Interactive} {Communication} ({RO}-{MAN})},
	author = {Tsamis, Georgios and Chantziaras, Georgios and Giakoumis, Dimitrios and Kostavelis, Ioannis and Kargakos, Andreas and Tsakiris, Athanasios and Tzovaras, Dimitrios},
	month = aug,
	year = {2021},
	pages = {520--526},
}

@article{neves_application_2020,
	title = {Application of mixed reality in robot manipulator programming},
	volume = {45},
	issn = {0143-991X},
	url = {https://doi.org/10.1108/IR-06-2018-0120},
	doi = {10.1108/IR-06-2018-0120},
	number = {6},
	urldate = {2024-09-28},
	journal = {Industrial Robot: An International Journal},
	publisher = {Emerald Publishing Limited},
	author = {Neves, João and Serrario, Diogo and Pires, J. Norberto},
	month = jan,
	year = {2020},
	pages = {784--793},
}

@article{ong_augmented_2020,
	title = {Augmented reality-assisted robot programming system for industrial applications},
	volume = {61},
	issn = {0736-5845},
	url = {https://www.sciencedirect.com/science/article/pii/S0736584519300250},
	doi = {10.1016/j.rcim.2019.101820},
	urldate = {2024-09-20},
	journal = {Robotics and Computer-Integrated Manufacturing},
	author = {Ong, S. K. and Yew, A. W. W. and Thanigaivel, N. K. and Nee, A. Y. C.},
	month = feb,
	year = {2020},
	pages = {101820},
}

@article{yang_har2bot_2024,
	title = {{HAR2bot}: a human-centered augmented reality robot programming method with the awareness of cognitive load},
	volume = {35},
	url = {https://www.scopus.com/inward/record.uri?eid=2-s2.0-85150410758&doi=10.1007%2fs10845-023-02096-2&partnerID=40&md5=42eb59880b346ff2a3d414038b270de6},
	doi = {10.1007/s10845-023-02096-2},
	number = {5},
	journal = {Journal of Intelligent Manufacturing},
	author = {Yang, Wenhao and Xiao, Qinqin and Zhang, Yunbo},
	year = {2024},
	pages = {1985 -- 2003},
}

@inproceedings{kumar_mixed_2025,
	title = {Mixed {Reality} {Outperforms} {Virtual} {Reality} for {Remote} {Error} {Resolution} in {Pick}-and-{Place} {Tasks}},
	url = {https://ieeexplore.ieee.org/abstract/document/10974002},
	doi = {10.1109/HRI61500.2025.10974002},
	urldate = {2025-12-03},
	booktitle = {2025 20th {ACM}/{IEEE} {International} {Conference} on {Human}-{Robot} {Interaction} ({HRI})},
	author = {Kumar, Advay and Simangunsong, Stephanie and Carreno-Medrano, Pamela and Cosgun, Akansel},
	month = mar,
	year = {2025},
	pages = {511--519},
}

@inproceedings{guo_large_2024,
	title = {Large {Language} {Model} {Based} {Multi}-agents: {A} {Survey} of {Progress} and {Challenges}},
	volume = {9},
	issn = {1045-0823},
	shorttitle = {Large {Language} {Model} {Based} {Multi}-agents},
	url = {https://www.ijcai.org/proceedings/2024/890},
	doi = {10.24963/ijcai.2024/890},
	language = {en},
	urldate = {2025-12-03},
	author = {Guo, Taicheng and Chen, Xiuying and Wang, Yaqi and Chang, Ruidi and Pei, Shichao and Chawla, Nitesh V. and Wiest, Olaf and Zhang, Xiangliang},
	month = aug,
	year = {2024},
	pages = {8048--8057},
}

\end{document}